\documentclass{article}

\usepackage{times}
\usepackage{graphicx} 
\usepackage{subfigure} 

\usepackage{natbib}

\usepackage{algorithm}
\usepackage{algorithmic}

\usepackage{hyperref}

\usepackage[accepted]{icml2015}

\icmltitlerunning{Modeling Order in Neural Word Embeddings at Scale}

\begin{document} 

\twocolumn[
\icmltitle{Modeling Order in Neural Word Embeddings at Scale}

\icmlauthor{Andrew Trask}{andrew.trask@digitalreasoning.com}
\icmladdress{Digital Reasoning Systems, Inc.,
            Nashville, TN USA}
\icmlauthor{David Gilmore}{david.gilmore@digitalreasoning.com}
\icmladdress{Digital Reasoning Systems, Inc.,
            Nashville, TN USA}
\icmlauthor{Matthew Russell}{matthew.russell@digitalreasoning.com}
\icmladdress{Digital Reasoning Systems, Inc.,
            Nashville, TN USA}

\icmlkeywords{boring formatting information, machine learning, ICML}

\vskip 0.3in
]

\begin{abstract} 
Natural Language Processing (NLP) systems commonly leverage bag-of-words co-occurrence techniques to capture semantic and syntactic word relationships. The resulting word-level distributed representations often ignore morphological information, though character-level embeddings have proven valuable to NLP tasks. We propose a new neural language model incorporating both word order and character order in its embedding. The model produces several vector spaces with meaningful substructure, as evidenced by its performance of 85.8\% on a recent word-analogy task, exceeding best published syntactic word-analogy scores by a 58\% error margin ~\cite{pennington-socher-manning:2014:EMNLP2014}. Furthermore, the model includes several parallel training methods, most notably allowing a skip-gram network with 160 billion parameters to be trained overnight on 3 multi-core CPUs, 14x larger than the previous largest neural network ~\cite{coates2013deep}.

\end{abstract} 

\section{Introduction}
\label{submission}

NLP systems seek to automate the extraction of useful information from sequences of symbols in human language. These systems encounter difficulty due to the complexity and sparsity in natural language. Traditional systems have represented words as atomic units with success in a variety of tasks~\cite{Katz1987}. This approach is limited by the curse of dimensionality and has been outperformed by neural network language models (NNLM) in a variety of tasks~\cite{Bengio:2003:NPL:944919.944966,morin2005hierarchical,NIPS2008_3583}. NNLMs overcome the curse of dimensionality by learning distributed representations for words~\cite{g.e.hintonj.l.mcclellandd.e.rumelhart1986,Bengio:2003:NPL:944919.944966}. Specifically, neural language models embed a vocabulary into a smaller dimensional linear space that models ``the probability function for word sequences, expressed in terms of these representations"~\cite{Bengio:2003:NPL:944919.944966}. The result is a vector space model \cite{maas2010probabilistic} that encodes semantic and syntactic relationships and has defined a new standard for feature generation in NLP~\cite{manning2008introduction,Sebastiani:2002:MLA:505282.505283,turian2010word}. 

NNLMs generate word embeddings by training a symbol prediction task over a moving local-context window such as predicting a word given its surrounding context~\cite{DBLP:journals/corr/abs-1301-3781,NIPS2013_5021}. This work follows from the distributional hypothesis: words that appear in similar contexts have similar meaning~\cite{harris54}. Words that appear in similar contexts will experience similar training examples, training outcomes, and converge to similar weights. The ordered set of weights associated with each word becomes that word's dense vector embedding. These distributed representations encode shades of meaning across their dimensions, allowing for two words to have multiple, real-valued relationships encoded in a single representation ~\cite{liang2015semantics}.

~\cite{conf/naacl/MikolovYZ13} introduced a new property of word embeddings based on word analogies such that vector operations between words mirror their semantic and syntactic relationships. The analogy ''king is to queen as man is to woman'' can be encoded in vector space by the equation king - queen ~= man - woman. A dataset of these analogies, the Google Analogy Dataset~\footnote{http://word2vec.googlecode.com/svn/trunk/}, is divided into two broad categories, semantic queries and syntactic queries. Semantic queries idenfity relationships such as ``France is to Paris as England is to London'' whereas syntactic queries identify relationships such as ``running is to run as pruning is to prune''. This is a standard by which distributed word embeddings may be evaluated.

Until recently, NNLMs have ignored morphology and word shape. However, including information about word structure in word representations has proven valuable for part of speech analysis~\cite{icml2014c2_santos14}, word similarity~\cite{luong2013better}, and information extraction~\cite{qi2014deep}. 

We propose a neural network architecture that explicitly encodes order in a sequence of symbols and use this architecture to embed both word-level and character-level representations. When these two representations are concatenated, the resulting representations exceed best published results in both the semantic and syntactic evaluations of the Google Analogy Dataset.

\section{Related Work} 
 
\subsection{Word-level Representations (Word2vec)}
Our technique is inspired by recent work in learning vector representations of words, phrases, and sentences using neural networks ~\cite{DBLP:journals/corr/abs-1301-3781,NIPS2013_5021,DBLP:journals/corr/LeM14}. In the CBOW configuration of the negative sampling training method by ~\cite{DBLP:journals/corr/abs-1301-3781}, each word is represented by a row-vector in matrix $syn_0$ and is concatenated, summed, or averaged with other word vectors in a context window. The resulting vector is used in a classifier $syn_1$ to predict the existence of the whole context with the the focus term (positive training) or absence of other randomly sampled words in the window (negative sampling). The scalar output is passed through a sigmoid function $(\sigma(z) = (1 + e^{(-z)})$, returning the network's probability that the removed word exists in the middle of the window, without stipulation on the order of the context words. This optimizes the following objective:

\[
\arg\max_\theta \prod_{(w,C)\in d} p(w=1|C;\theta) \prod_{(w,C)\in d'} p(w=0|C;\theta) 
\]

where $d$ represents the document as a collection of context-word pairs $(w,C)$ and $C$ is an unordered group of words in a context window. $d'$ is a set of random $(w,C)$ pairs. $\theta$ will be adjusted such that $p(w=1,C;\theta)$ = $1$ for context-word pairs that exist in $d$, and $0$ for random context-word pairs that do not exist in $d'$. In the skip-gram negative sampling work by ~\cite{DBLP:journals/corr/abs-1301-3781,NIPS2013_5021}, each word in a context is trained in succession. This optimizes the following objective:

\[
\arg\max_\theta \prod_{(w,c)\in d} p(w=1|c;\theta) \prod_{(w,c)\in d'} p(w=0|c;\theta) 
\]

where $d$ represents the document as a collection of context-word pairs $(w,c)$ and $c$ represents a single word in the context. Modeling an element-wise probability that a word occurs given another word in the context, the element-wise nature of this probability allows (2) to be an equivalent objective to the skip-gram objective outlined in~\cite{NIPS2013_5021,DBLP:journals/corr/GoldbergL14}.

Reducing the window size under these models constrains the probabilities to be more localized, as the probability that two words co-occur will reduce when the window reduces which can be advantageous for words subject to short-windowed statistical significance. For example, currency symbols often co-occur with numbers within a small window. Outside of a small window, currency symbols and numbers are not likely to co-occur. Thus, reducing the window size reduces noise in the prediction. Words such as city names, however, prefer wider windows to encode broader co-occurrence statistics with other words such as landmarks, street-names, and cultural words which could be farther away in the document.

\begin{figure}[ht]
\vskip 0.0in
\begin{center}
\centerline{\includegraphics[width=\columnwidth]{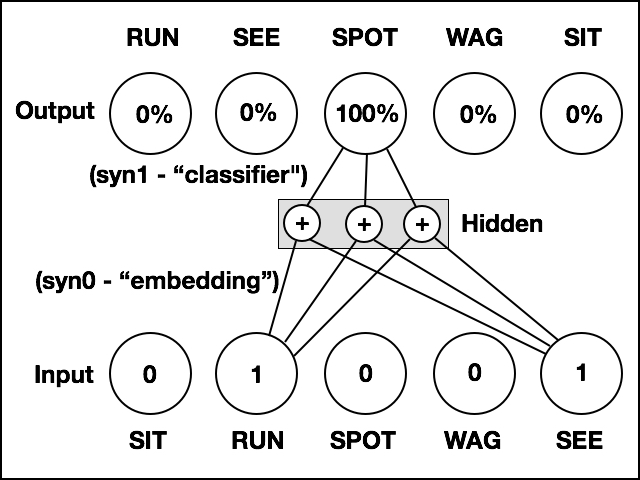}}
\caption{Diagram of word2vec's Continuous Bag of Words training method over the sentence ``SEE SPOT RUN". Embeddings for ``SEE" and ``RUN" are summed into a third vector that is used to predict the probability that the middle word is ``SPOT". }
\label{icml-historical}
\end{center}
\vskip -0.3in
\end{figure} 

Neither skip-gram nor CBOW explicitly preserve word order in their word embeddings~\cite{DBLP:journals/corr/abs-1301-3781,NIPS2013_5021,DBLP:journals/corr/LeM14}. Ordered concatenation of $syn_0$ vectors does embed order in $syn_1$, but this is obfuscated by the fact that the same embedding for each word must be linearly compatible with the feature detectors in every window position. In addition to changing the objective function, this has the effect of cancelling out features that are unique to only one window position by those in other window positions that are attempting to be encoded in the same feature detector dimension. This effect prevents word embeddings from preserving order based features. The other methods (sum, average, and skip-gram) ignore all order completely in their modeling and model only co-occurrence based probability in their embeddings.

\subsection{Character-level Representations}

Recent work has explored techniques to embed word shape and morphology features into word embeddings. The resulting embeddings have proven useful for a variety of NLP tasks.

\subsubsection{Deep Neural Network}

~\cite{icml2014c2_santos14} proposed a Deep Neural Network (DNN) that ``learns character-level representation[s]
of words and associate[s] them with usual word representations to perform POS tagging." The resulting embeddings were used to produce state-of-the-art POS taggers for both English and Portuguese data sets. The network architecture leverages the convolutional approach introduced in \cite{Waibel:1990:PRU:108235.108263} to produce local features around each character of the word and then combines them into a fixed-sized character-level embedding of the word. The character-level word embedding is then concatenated with a word-level embedding learned using word2vec. Using only these embeddings, ~\cite{icml2014c2_santos14} achieves state-of-the-art results in POS tagging without the use of hand-engineered features.

\subsubsection{Recursive Neural Network}

\cite{luong2013better} proposed a ``novel model that is capable of building representations for morphologically complex words from their morphemes." The model leverages a recursive neural network (RNN)~\cite{socher2011parsing} to model morphology in a word embedding. Words are decomposed into morphemes using a morphological segmenter~\cite{creutz2007unsupervised}. Using the ``morphemic vectors", word-level representations are constructed for complex words. In the experiments performed by \cite{luong2013better}, word embeddings were borrowed from \cite{huang2012improving} and \cite{DBLP:journals/corr/abs-1103-0398}. After conducting a morphemic segmentation, complex words were then enhanced with morphological feature embeddings by using the morphemic vectors in the RNN to compute word representations ``on the fly". The resulting model outperforms existing embeddings on word similarity tasks accross several data sets.

\section{The Partitioned Embedding Neural Network Model (PENN)} 

\begin{figure}[ht]
\vskip -0in
\begin{center}
\centerline{\includegraphics[width=\columnwidth]{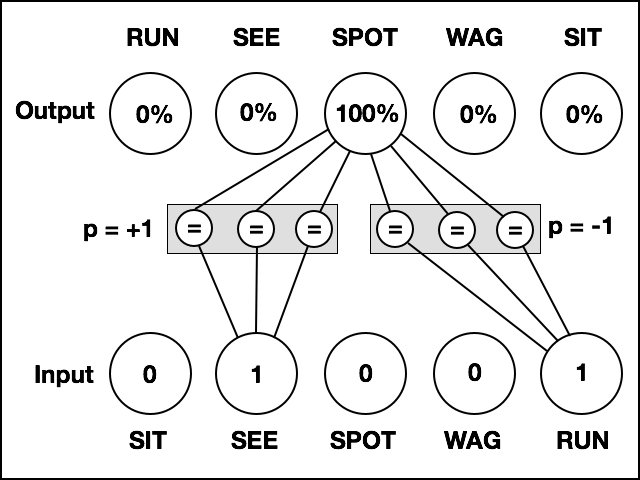}}
\caption{The Windowed configuration of PENN when using the CLOW training method modeling ``SEE SPOT RUN". }
\label{icml-historical}
\end{center}
\vskip -0.3in
\end{figure} 

We propose a new neural language model called a Partitioned Embedding Neural Network (PENN). PENN improves upon word2vec by modeling the order in which words occur. It models order by partitioning both the embedding and classifier layers. There are two styles of training corresponding to the CBOW negative sampling and skip-gram negative sampling methods in word2vec, although they differ in key areas. 

	The first property of PENN is that each word embedding is partitioned. Each partition is trained differently from each other partition based on word order, such that each partition models a different probability distribution. These different probability distributions model different perspectives on the same word. The second property of PENN is that the classifier has different inputs for words from different window positions. The classifier is partitioned with equal partition dimensionality as the embedding. It is possible to have fewer partitions in the classifier than the embedding, such that a greater number of word embeddings are summed/averaged into fewer classifier partitions. This configuration has better performance when using smaller dimensionality feature vectors with large windows as it balances the (embedding partition size) / (window size) ratio. The following subsection presents the two opposite configurations under the PENN framework.

\subsection{Plausible Configurations}

\subsubsection{Windowed}

The simplest configuration of a PENN architecture is the \textit{windowed} configuration, where each partition corresponds to a unique window position in which a word occurs. As illustrated in Figure 2, if there are two window positions (one on each side of the focus term), then each embedding would have two partitions. When a word is in partition p = +1 (the word before the focus term), the partition corresponding to that position is propagated forward, and subsequently back propagated into, with the p = -1 partition remaining unchanged. 

\subsubsection{Directional}

The opposite configuration to windowed PENN is the \textit{directional} configuration. Instead of each partition corresponding to a window position, there are only two partitions. One partition corresponds to every positive, forward predicting window position (left of the focus term) and the other partition corresponds to every negative, backward predicting window position (right of the focus term). For each partition, all embeddings corresponding to that partition are summed or averaged when being propagated forward.

\begin{figure}[H]
\vskip 0in
\begin{center}
\centerline{\includegraphics[width=\columnwidth]{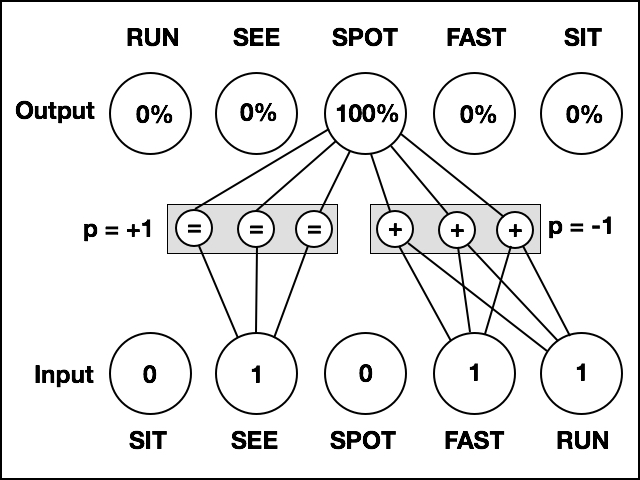}}
\caption{The Directional configuration of PENN when using the CLOW training method. It is modeling the sentence ``SEE SPOT RUN FAST". }
\label{icml-historical}
\end{center}
\vskip -0.2in
\end{figure} 

\subsection{Training Styles}

\subsubsection{Continuous List of Words (CLOW)}

The Continuous List of Words (CLOW) training style under the PENN framework optimizes the following objective function:

\begin{center}
\[
\arg\max_\theta (\prod_{(w,C)\in d}\prod_{-c\leq j\leq c, j\neq 0} p(w=1|c_j^j;\theta) 
\]
\end{center}

\begin{center}
\[
\prod_{(w,C)\in d'}\prod_{-c\leq j\leq c, j\neq 0} p(w=0|c_j^j;\theta))
\]
\end{center}

where $c_j^j$ is the \textit{location specific representation} (partition $^j$) for the word at window position $j$ relative to the focus word $w$. Closely related to the CBOW training method, the CLOW method models the probability that in an ordered list of words, a specific word is present in the middle of the list, given the presence and location of the other words. For each training example out of a windowed sequence of words, the middle “focus” term is removed. Then, a partition is selected from each remaining word's embedding based on that word's position relative to the focus term. These partitions are concatenated and propagated through the classifier layer. All weights are updated to model the probability that the presence of the focus term is 100\% (positive training) and other randomly sampled words 0\% (negative sampling).

\subsubsection{Skip-Gram}

The skip-gram training style under the PENN framework optimizes the following objective function

\begin{center}
\[
\arg\max_\theta (\prod_{(w,C)\in d}\sum_{-c\leq j\leq c, j\neq 0} p(w_j=1|c_j^j;\theta)
\]

\end{center}

\begin{center}
\[
\prod_{(w,C)\in d'}\sum_{-c\leq j\leq c, j\neq 0} p(w_j=0|c_j^j;\theta)) 
\]

\end{center}
where, like CLOW, $c_j^j$ is the \textit{location specific representation} (partition $^j$) for the word at window position $j$ relative to the focus word $w$. $w_j$ is the relative location specific probability (partition) of the focus term. PENN skip-gram is almost identical to the CLOW method with one key difference. Instead of each partition of a word being concatenated with partitions from neighboring words, each partition is fed forward and back propagated in isolation. This models the probability that, given a single word, the “focus” term is present a relative number of words away in a given direction. This captures information lost in the word2vec skip-gram architecture by modeling based on the relative location of a context word in the window as opposed to an arbitrary location within the window.

The intuition behind modeling $w$ and $c$ based on $j$ at the same time becomes clear when considering the neural architecture of these embeddings. Partitioning the context word into $j$ partitions gives a \textit{location specific representation} for a word's relative position. Location specific representations are important even for words with singular meanings. Consider the word ``going", a word of singular meaning. This word's effect on a task predicting a word immediately before it is completely different than predicting a word immediately after it. The phrase ``am going" is a plausible phrase. The phrase ``going am" is not. Thus, forcing this word to have a consistent embedding across these tasks forces it to convey identical information optimizing for nonidentical problems.

Partitioning the classifier incorporates this same principle with respect to the focus word. The focus word will read features presented to it in a different light with a different weighting given its position. For example, ``dollars" is far more likely to be predicted accurately based on the word before it; whereas, it is not likely to be predicted correctly by a word ten window positions after. Thus, the classifier responsible for looking for features indicating that ``dollars" is next should not have to be the same classifier that looks for features ten window positions into the future. Training separate classifier partitions based on window position avoids this phenomenon.

\subsection{Distributed Training Optimizations}

\subsubsection{Skip-Gram}

When skip-gram is used to model ordered sets of words under the PENN framework each classifier partition and its associated embedding partitions may be trained in full-parallel (with no inter-communication) and reach the exact same state as if they were not distributed. A special case of this is the \textit{windowed} embedding configuration, where every window position can be trained in full parallel and concatenated (embeddings and classifiers) at the end of training. This allows very large, rich embeddings to be trained on relatively small, inexpensive machines in a small amount of time with each machine optimizing a part of the overall objective function. Given machine $j$, training skip-gram under the \textit{windowed} embedding configuration optimizes the following objective function:

\begin{center}
\[
\arg\max_\theta (\prod_{(w,C)\in d} p(w_j=1|c_j^j;\theta) 
\]
\end{center}
\begin{center}
\[
\prod_{(w,C)\in d'}p(w_j=0|c_j^j;\theta)) 
\]
\end{center}

Concatenation of the weight matrices $syn_0$ and $syn_1$ then incorporates the sum over $j$ back into the PENN skip-gram objective function during the forward propagation process, yielding identical training results as a network trained in a single-threaded, single-model PENN skip-gram fashion. This training style achieves parity training results with current state-of-the-art methods while training in parallel over as many as $j$ separate machines.

\subsubsection{CLOW}
The CLOW method is an excellent candidate for the ALOPEX distributed training algorithm \cite{unnikrishnan1994alopex} because it trains on very few (often single) output probabilities at a time. Different classifier partitions may be trained on different machines, with each training example sending a short list of floats per machine across the network. They all share the same global error and continue on to the next iteration.

A second, nontrivial optimization is found in the strong performance of the \textit{directional} CLOW implementation with very small window sizes (pictured below with a window size of 1). \textit{Directional} CLOW is able to achieve a parity score using a window size of 1, contrasted with word2vec using a window size of 10 when all other parameters are equal, reducing the overall training time by a factor of 10.

\section{Dense Interpolated Embedding Model}

\begin{table}[h]
\centering
\begin{tabular}{|c|c|c|c|}
\hline
\multicolumn{4}{|c|}{\textbf{char similarity}} \\ \hline
\textbf{a} & \textbf{A} & \textbf{1} & \textbf{s} \\ \hline
o & E & 5 & p \\
e & O & 7 & h \\
i & I & 4 & x \\
u & ! & 8 & d \\ \hline
\end{tabular}
\caption{A focus character and the 4 closest characteres ordered by cosine similarity.}
\label{my-label}
\end{table}

\begin{table}[h]
\centering
\begin{tabular}{|c|c|c|c|}
\hline
\multicolumn{2}{|c|}{\textbf{SEMANTIC}} & \multicolumn{2}{c|}{\textbf{SYNTACTIC}} \\ \hline
\multicolumn{4}{|c|}{\textbf{``general'' - similarity}}                             \\ \hline
secretary               & 0.619         & gneral                & 0.986           \\
elections               & 0.563         & genral                & 0.978           \\
motors                  & 0.535         & generally             & 0.954           \\
undersecretary          & 0.534         & generation            & 0.944           \\ \hline
\multicolumn{4}{|c|}{\textbf{``sees" - ``see" + ``bank" \~=}}                         \\ \hline
firestone               & 0.580         & banks                 & 0.970           \\
yard                    & 0.545         & bank                  & 0.939           \\
peres                   & 0.506         & balks                 & 0.914           \\
c.c                     & 0.500         & bans                  & 0.895           \\ \hline
\end{tabular}
\caption{An example of syntactic vs semantic embeddings on the cosine similarity and word-analogy tasks.}
\label{my-label}
\end{table}

We propose a second new neural language model called a Dense Interpolated Embedding Model (DIEM). DIEM uses neural embeddings learned at the character level to generate a fixed-length syntactic embedding at the world level useful for syntactic word-analogy tasks, leveraging patterns in the characters as a human might when detecting syntactic features such as plurality.

\subsection{Method}

\begin{algorithm}[h]
   \caption{Dense Interpolated Embedding Pseudocode}
   \label{alg:example}
\begin{algorithmic}
   \STATE {\bfseries Input:} wordlength $I$, list char embeddings (e.g. the word) $char_i$, multiple $M$, char dim $C$, vector $v_m$
   \FOR{$i=0$ {\bfseries to} $I-1$}
   \STATE $s = $M * $i / $l
   \FOR{$m=0$ {\bfseries to} $M-1$}
   \STATE $d$ = $pow$(1 - ($abs$($s$ - $m$)) / $M$,2)
   \STATE $v_m$ = $v_m$ + $d$ * $char_{i}$
   \ENDFOR
   \ENDFOR
\end{algorithmic}
\vskip -0.0in
\end{algorithm}

Generating syntactic embeddings begins by generating character embeddings. Character embeddings are generated using vanilla word2vec by predicting a focus character given its context. This clusters characters in an intuitive way, vowels with vowels, numbers with numbers, and capitals with capitals. In this way, character embeddings represent morphological building blocks that are more or less similar to each other, based on how they have been used.

Once character embeddings have been generated, interpolation may begin over a word of length $I$. The final embedding size must be selected as a multiple $M$ of the character embedding dimensionality $C$. For each character in a word, its index $i$ is first scaled linearly with the size of the final ``syntactic" embedding such that $s$ = $M$ * $i$ / $l$. Then, for each length $C$ position $m$ (out of $M$ positions) in the final word embedding $v_m$, a squared distance is calculated relative to the scaled index such that distance $d$ = $pow$(1-(abs($s$ - $j$)) / $M$,2). The character vector for the character at position $i$ in the word is then scaled by $d$ and added elementwise into position $m$ of vector $v$.

A more efficient form of this process caches a set of transformation matrices, which are cached values of $d_{i,m}$ for words of varying size. These matrices are used to transform variable length concatenated character vectors into fixed length word embeddings via vector-matrix multiplication.

These embeddings are useful for a variety of tasks, including syntactic word-analogy queries. Furthermore, they are useful for syntactic query expansion, mapping sparse edge cases of a word (typos, odd capitalization, etc.) to a more common word and its semantic embedding. 

\subsection{Distributed Use and Storage Optimizations}

Syntactic vectors also provide significant scaling and generalization advantages over semantic vectors. New syntactic vectors may be inexpensively generated for words never before seen, giving loss-less generalization to any word from initial character training, assuming only that the word is made up of characters that have been seen. Syntactic embeddings can be generated in a fully distributed fashion and only require a small vector concatenation and vector-matrix multiplication per word. Secondly, the character vectors (typically length 32) and transformation matrices (at most 20 or so of them) can be stored very efficiently relative to the semantic vocabularies, which can be several million vectors of dimensionality 1000 or more. Despite their significant positive impact on quality, DIEM optimally performs using 6+ orders of magnitude less storage space, and 5+ orders of magnitude fewer training examples than word-level semantic embeddings.

\section {Experiments}

\begin{figure*}
  \includegraphics[width=\textwidth]{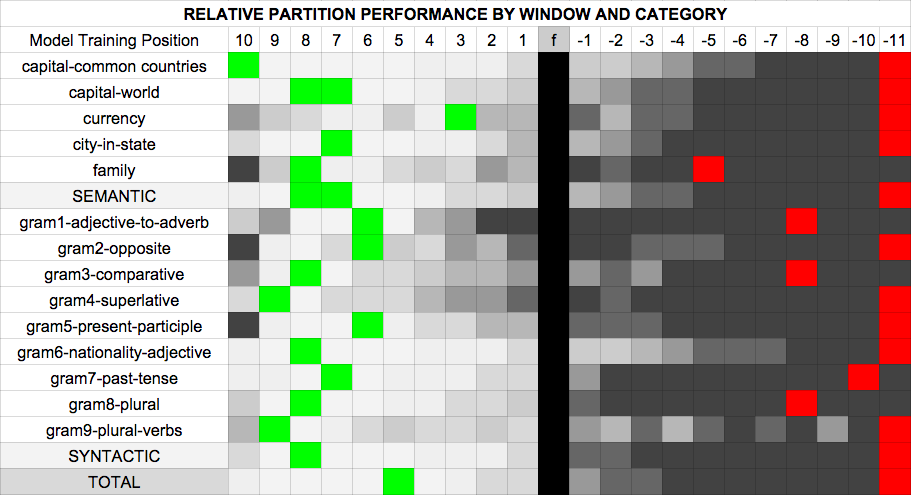}
  \caption{Green represents the highest quality partition. Red indicates the lowest. Gray indicates the gradient performance between red and green. Two greens in the same row indicates a tie within a 1\% margin.}
\end{figure*}
\subsection{Evaluation Methods}

We conduct experiments on the word-analogy task of ~\cite{DBLP:journals/corr/abs-1301-3781}. It is made up of a variety of word similarity tasks, as described in (Luong et al., 2013). Known as the “Google Analogy Dataset”, it contains 19,544 questions asking ''a is to b as c is to \_'' and is split into semantic and syntactic sections. Both sections are further divided into subcategories based on analogy type, as indicated in the results tables below.

All training occurs over the dataset available from the Google word2vec website\footnote{https://code.google.com/p/word2vec/}, using the packaged word-analogy evaluation script. The dataset contains approximately 8 billion words collected from English News Crawl, 1-Billion-Word Benchmark, UMBC Webbase, and English Wikipedia. The dataset used leverages the default data-phrase2.txt normalization in all training, which includes both single tokens and phrases. Unless otherwise specified, all parameters for training and evaluating are identical to the default parameters specified in the default word2vec “big” model, which is freely available online.
	
\subsection{Embedding Partition Relative Evaluation}

Figure 4 displays the relative accuracy of each partition in a PENN model as judged by row-relative word-analogy scores. Other experiments indicated that the pattern present in the heat-map is consistent across parameter tunings. There is a clear quality difference between window positions that predict forward (left side of the figure) and window positions that predict backward (right side of the figure). ``currency'' achieves most of its predictive power in short range predictions, whereas ``capital-common countries'' is a much smoother gradient over the window. These patterns support the intuition that different window positions play different roles in different tasks.

\subsection{Evaluation of CLOW and CBOW}

\begin{table}[h]
\begin{tabular}{|l|ccc|}
\hline
Configuration Style   & W2V   & Window         & (see tbl. 5)         \\ \hline
Training Style        & CBOW  & CLOW           & ENSEM          \\ \hline
Word Vector Size      & \textbf{2000}  & \textbf{2000}  & \textbf{7820} \\
Partition Size        & 2000  & 500           & (see tbl. 5)    \\
Window Size           & 10    & 2             & (see tbl. 5)    \\ \hline
capital-common        & 85.18 & \textbf{98.81}          & 95.65          \\
capital-world         & 75.38 & 90.01          & \textbf{93.90}          \\
currency              & 0.40  & 16.89          & \textbf{17.32}          \\
city-in-state         & 65.18 & 78.31          & \textbf{78.88}          \\
family                & 49.01 & 84.39          & \textbf{85.35}          \\ \hline
SEMANTIC              & 65.11 & 80.62          & \textbf{82.70}          \\ \hline
adjective-to-adverb   & 15.62 & 30.04          & \textbf{90.73}          \\
opposite              & 8.50  & 38.55          & \textbf{73.15}          \\
comparative           & 51.95 & 94.37          & \textbf{99.70}          \\
superlative           & 33.87 & 79.77          & \textbf{91.89}          \\
present-participle    & 45.45 & 81.82          & \textbf{93.66}          \\
nationality-adjective & 88.56 & 89.38          & \textbf{91.43}          \\
past-tense            & 55.19 & \textbf{76.99}          & 60.01          \\
plural                & 73.05 & 83.93          & \textbf{97.90}          \\
plural-verbs          & 28.74 & 73.33          & \textbf{95.86}          \\ \hline
SYNTACTIC             & 49.42 & 75.11          & \textbf{88.29} \\ \hline
TOTAL                 & 56.49 & 77.59          & \textbf{85.77} \\ \hline
\end{tabular}
\caption{\centering
Comparison between Word2vec, CLOW, and Penn-DIEM Ensemble}
\end{table}

Table 3 shows the performance of the default CBOW implementation of word2vec relative to CLOW and DIEM when configured to 2000 dimensional embeddings. Between tables 3 and 4, we see that increasing dimensionality of baseline CBOW word2vec past 500 achieves sub-optimal performance. Thus, a fair comparison of two models should be between optimal (as opposed to just identical) parameterization for each model. This is especially important given that PENN models are modeling a much richer probability distribution, given that order is being preserved. Thus, optimal parameter settings often require larger dimensionality.
Unlike the original CBOW word2vec, we have found that bigger window size is not always better. Larger windows tend to create slightly more semantic embeddings, whereas smaller window sizes tend to create slightly more syntactic embeddings. This follows the intuition that syntax plays a huge role in grammar, which is dictated by rules about which words make sense to occur immediately next to each other. Words that are +5 words apart cluster based on subject matter and semantics as opposed to grammar. With respect to window size and overall quality, because partitions slice up the global vector for a word, increasing the window size decreases the size of each partition in the window if the global vector size remains constant. Since each embedding is attempting to model a very complex (hundreds of thousands of words) probability distribution, the partition size in each partition must remain high enough to model this distribution. Thus, modeling large windows for semantic embeddings is optimal when using either the \textit{directional} embedding model, which has a fixed partition size of 2, or a large global vector size. The \textit{directional} model with optimal parameters has slightly less quality than the \textit{windowed} model with optimal parameters due to the vector averaging occurring in each window pane.

\subsection{Evaluation of DIEM Syntactic Vectors on Syntactic Tasks}

\begin{table}[H]
\vskip -0.1in
\begin{tabular}{|l|ll|l|}
\hline
Semantic Architecture & \multicolumn{1}{l|}{CBOW} & CLOW  & DIEM    \\ \hline
Semantic Vector Dim.  & 500                       & 500   & 500            \\ \hline
SEMANTIC TOTAL        & 81.02                     & 80.19 & 80.19          \\ \hline
adjective-to-adverb   & 37.70                     & 35.08 & \textbf{94.55} \\
opposite              & 36.21                     & 40.15 & \textbf{74.60} \\
comparative           & 86.71                     & 87.31 & \textbf{92.49} \\
superlative           & 80.12                     & 82.00 & \textbf{87.61} \\
present-participle    & 77.27                     & 80.78 & \textbf{93.27} \\
nationality-adjective & \textbf{90.43}                     & 90.18 & 71.04          \\
past-tense            & 72.37                     & \textbf{73.40} & 47.56          \\
plural                & 80.18                     & 81.83 & \textbf{93.69} \\
plural-verbs          & 58.51                     & 63.68 & \textbf{95.97} \\ \hline
SYNTACTIC TOTAL       & 72.04                     & 73.45 & \textbf{81.53} \\ \hline
COMBINED SCORE        & 76.08                     & 76.49 & \textbf{80.93} \\ \hline
\end{tabular}
\caption{Above we see can observe the boost that syntactic based DIEM feature vectors gives our unsupervised semantic models, relative to both word2vec-CBOW and CLOW}
\label{my-label}
\vskip -0.0in
\end{table}

\begin{table}[h]
\begin{tabular}{|c|c|c|}
\hline
Conf. Training Style & Window Size & Dimensionality \\ \hline
Windowed           & 10          & 500            \\
Directional           & 5           & 500            \\
Windowed           & 2           & 2000           \\
Directional           & 5           & 2000           \\
Directional           & 10          & 2000           \\
Directional           & 1           & 500            \\
DIEM           & x          & 320           \\\hline
\end{tabular}
\caption{\centering
Concatenated Model Configurations}
\end{table}

Table 4 documents the change in syntactic analogy query quality as a result of the interpolated DIEM vectors. For the DIEM experiment, each analogy query was first performed by running the query on CLOW and DIEM independently, and selecting the top thousand CLOW cosine similarities. We summed the squared cosine similarity of each of these top thousand with each associated cosine similarity returned by the DIEM and resorted. This was found to be an efficient estimation of concatenation that did not reduce quality.

Table 5 documents the parameter selection for a combined neural network partitioned according to several training styles and dimensionalities. As in the experiments of Table 3, each analogy query was first performed by running the query on each model independently, selecting the top thousand cosine similarities. We summed the cosine similarity of each of these top thousand entries across all models (excluding DIEM for semantic queries) and resorted. (For normalization purposes, DIEM scores were raised to the power of 10 and all other scores were raised to the power of 0.1 before summing).

\subsection{High Level Comparisons}

\begin{table}[H]
\begin{tabular}{|c|c|cc|cc|}
\hline
\textbf{Algorithm} & \textbf{GloVe} & \multicolumn{2}{c|}{\textbf{Word2Vec}} & \multicolumn{2}{c|}{\textbf{PENN+D}} \\ \hline
Config             &  x              & CBOW               & SG                & SG             & ENS          \\
Params             & x              & 7.6 B              & 7.6 B             & \textbf{40B}    & \textbf{59B}    \\
Sem. Dims          & 300            & 500                & 500               & 5000             & 7820            \\
Semantic           & 81.9           & 81.0               & 82.2              & 69.6             & \textbf{82.7}   \\
Syntactic          & 69.3           & 72.0               & 71.3              & 80.0             & \textbf{88.3}   \\
Combined           & 75.0           & 76.1               & 76.2              & 75.3             & \textbf{85.8}   \\ \hline
\end{tabular}
\caption{Scores reflect best published results in each category, semantic, syntactic, and combined when parameters are tuned optimally for each individual category.}
\vskip -0.2in
\end{table}

Our final results show a lift in quality and size over previous models with a 58\% syntactic lift over the best published syntactic result, and a 40\% overall lift over the best published overall result \cite{pennington-socher-manning:2014:EMNLP2014}. Table 5 also includes the highest word2vec scores we could achieve through better parameterization (which also exceeds the best published word2vec scores). Within PENN models, there exists a speed vs. performance tradeoff between SG-DIEM and CLOW-DIEM. In this case, we achieve a 20x level of parallelism in SG-DIEM relative to CLOW, with each model training partitions of 250 dimensions (250 * 20 = 5000 final dimensionality). A 160 billion parameter network was also trained overnight on 3 multi-core CPUs, however it yielded 20000 dimensional vectors for each word and subsequently overfit the training data. This is because a dataset of 8 billion tokens with a negative sampling parameter of 10 has 80 billion training examples. Having more parameters than training examples overfits a dataset, whereas 40 billion performs at parity with current state of the art, as pictured in Table 5. Future work will experiment with larger datasets and vocabularies. The previous largest neural network contained 11.2 billion parameters~\cite{coates2013deep}, whereas CLOW and the largest SG contain 16 billion (trained all together) and 160 billion (trained across a cluster) parameters respectively as measured by the number of weights.

\section {Conclusion and Future Work}

Encoding both word and character order in neural word embeddings is beneficial for word-analogy tasks, particularly syntactic tasks. These findings are based upon the intuition that order matters in human language and has been validated through the methods above. Future work will further investigate the scalability of these word embeddings to larger datasets with reduced runtimes.

\nocite{langley00}

\bibliography{example_paper}
\bibliographystyle{icml2015}

\end{document}